\documentclass[manuscript]{acmart}
\AtBeginDocument{%
  \providecommand\BibTeX{{%
    \normalfont B\kern-0.5em{\scshape i\kern-0.25em b}\kern-0.8em\TeX}}}

\copyrightyear{2022}
\acmYear{2022}
\setcopyright{acmlicensed}\acmConference[FAccT '22]{2022 ACM Conference on Fairness, Accountability, and Transparency}{June 21--24, 2022}{Seoul, Republic of Korea}
\acmBooktitle{2022 ACM Conference on Fairness, Accountability, and Transparency (FAccT '22), June 21--24, 2022, Seoul, Republic of Korea}
\acmPrice{15.00}
\acmDOI{10.1145/3531146.3533170}
\acmISBN{978-1-4503-9352-2/22/06}

\usepackage{amsmath}
\usepackage{amsthm}
\usepackage{graphicx}
\usepackage{natbib}
\usepackage[ruled,vlined]{algorithm2e}
\usepackage{url} 
\usepackage[shortlabels]{enumitem}
\usepackage[ruled]{algorithm2e}
\usepackage{subcaption}
\usepackage{bm}

\DeclareMathOperator*{\argmax}{argmax}
\DeclareMathOperator*{\argmin}{argmin}
\usepackage[font={small}]{caption}
\usepackage{xcolor}
\definecolor{linkcolor}{rgb}{0,0.2,0.6}
\DeclareMathOperator{\dep}{\not\! \perp\!\!\!\perp}

\begin{document}
\theoremstyle{plain}
\newtheorem{defn}{Definition}

\title{Rational Shapley Values}

\author{David S. Watson}
\orcid{0000-0001-9632-2159}
\affiliation{%
  \institution{Department of Statistical Science, University College London}
  \city{London}
  \country{UK}
}
\email{david.watson@ucl.ac.uk}

\begin{abstract}
  Explaining the predictions of opaque machine learning algorithms is an important and challenging task, especially as complex models are increasingly used to assist in high-stakes decisions such as those arising in healthcare and finance. Most popular tools for post-hoc explainable artificial intelligence (XAI) are either insensitive to context (e.g., feature attributions) or difficult to summarize (e.g., counterfactuals). In this paper, I introduce \emph{rational Shapley values}, a novel XAI method that synthesizes and extends these seemingly incompatible approaches in a rigorous, flexible manner. I leverage tools from decision theory and causal modeling to formalize and implement a pragmatic approach that resolves a number of known challenges in XAI. By pairing the distribution of random variables with the appropriate reference class for a given explanation task, I illustrate through theory and experiments how user goals and knowledge can inform and constrain the solution set in an iterative fashion. The method compares favorably to state of the art XAI tools in a range of quantitative and qualitative comparisons.  
\end{abstract}




\maketitle

\section{Introduction}
\label{intro}
Machine learning algorithms are increasingly used in a variety of high-stakes domains, from credit scoring to medical diagnosis. However, many of the most popular and powerful models are \emph{opaque}, insomuch as end users cannot understand the reasoning that goes into individual predictions. A subdiscipline of computational statistics has rapidly emerged to address this issue, variously referred to as \emph{interpretable machine learning} or \emph{explainable artificial intelligence} (XAI). For good introductory surveys, see, e.g., \citep{Adadi2018, Guidotti_2018, Murdoch2019}. 

The most prominent methods in this burgeoning area are post-hoc, model-agnostic tools such as feature attributions and counterfactuals. Both approaches have recently come under fire for failing to properly handle dependencies between covariates. Most algorithms for computing Shapley values -- a popular feature attribution technique, formally defined in Sect.~\ref{sec:background} -- implicitly treat predictors as mutually independent, assigning positive weight to potentially impossible data permutations \citep{frye2019asymmetric, kumar2020}. Much recent work in counterfactual explanation has focused on how to model the impact of interventions on downstream variables, a reformulation that requires an explicitly causal approach \citep{mahajan2019, karimi2021}.

Others have pushed back against these critiques. \citet{Sundararajan2019} demonstrate that conditioning on covariates can misleadingly assign importance to irrelevant features. \citet{Janzing2020} point out that standard supervised learning algorithms do not explicitly model dependencies between features, and so intervening to set predictors to some fixed value is not just permissible but proper when computing Shapley values. 

Often lost in these debates is the seemingly obvious realization that the “right” explanation depends crucially on who is asking and why. Current explainability methods tend to treat explanations as objective deliverables that can be computed from just two ingredients: a target function and an accompanying dataset of real or simulated samples. This ignores a basic fact long acknowledged in philosophy and social science -- that \emph{explanations are fundamentally pragmatic}. To succeed, they must be tailored to the inquiring agent, who requests an explanation on the basis of certain beliefs and interests. 

If pragmatic approaches are few and far between in the XAI literature, perhaps this is because it is not obvious how we should formalize abstract, subjective notions like beliefs and interests. In this paper, I use tools from Bayesian decision theory to do just that. I make three main contributions: (1) I combine disparate work in feature attributions and counterfactuals, demonstrating how the two can join forces to create more useful and flexible explanations. (2) I extend the axiomatic guarantees of Shapley values, allowing user beliefs and preferences to guide the search for a relevant subspace of contrastive baselines. (3) I implement an expected utility maximization procedure that optionally incorporates causal information to compute optimal attribution sets for individual agents. 

The remainder of this paper is structured as follows. In Sect.~\ref{sec:background}, I review important background material in XAI, with a focus on Shapley values and counterfactuals, as well as the essential formalisms of decision theory and structural causal models. In Sect.~\ref{sec:method}, I compare several different reference distributions, propose a new desideratum for Shapley values, and introduce an algorithm for computing them. In Sect.~\ref{sec:exp}, I put this method to work on a number of benchmark datasets. Following a discussion in Sect.~\ref{sec:discussion}, Sect.~\ref{sec:conclusion} concludes.

\section{Background}\label{sec:background}

In what follows, I use uppercase italics to represent variables, e.g. $X$; lowercase italics to represent their values, e.g. $x$; uppercase boldface to represent matrices, e.g. $\bm{X}$; lowercase boldface to represent vectors, e.g. $\bm{x}$; and calligraphic type to represent the support of a distribution, e.g. $\mathcal{X}$.

\subsection{XAI Methods}

Originally developed in the context of cooperative game theory \citep{Shapley1953}, Shapley values have been adapted for model interpretability by numerous authors. They represent the contribution of each feature toward a particular prediction. Let $\bm{x}_i \in \mathcal{X} \subseteq \mathbb{R}^d$ denote an input datapoint and $f(\bm{x}_i) = \hat{y}_i \in \mathcal{Y} \subseteq \mathbb{R}$ the corresponding output of our target function $f$ for each sample $i \in [n] = \{1, \dots, n\}$.\footnote{The model $f$ may be a classifier, although in this case we typically require class probabilities and work on the logit scale. For ease of exposition, in this section we will assume $f$ is a regressor.} Shapley values express this value as a sum: 
\begin{equation} \label{eq:additive}
    f(\bm{x}_i) = \sum_{j=0}^{d}\phi_j(i),
\end{equation}
where $\phi_0(i)$ represents $i$'s ``baseline'' (i.e., the default output of $f$; more on this shortly) and $\phi_j(i)$ the weight assigned to feature $X_j$ at point $\bm{x}_i$. Feature attribution methods that satisfy Eq.~\ref{eq:additive} are said to be \emph{efficient}. Let $v: [n] \times 2^d \rightarrow \mathbb{R}$ be a value function such that $v(i, S)$ is the payoff associated with feature subset $S \subseteq [d]$ and $v(i, \{\emptyset\}) = 0$ for all $i \in [n]$. The Shapley value $\phi_j(i)$ is given by a weighted average of $j$'s marginal contributions to all subsets that exclude it:
\begin{equation} \label{eq:shapley}
    \phi_j(i) = \frac{1}{d!} \sum_{S \subseteq [d] \backslash \{j\}} |S|!(d - |S| - 1)!~v(i, S \cup \{j\}) - v(i, S).
\end{equation}
Several properties of this value are worth noting.
\begin{itemize}
    \item \emph{Linearity}. For all $j \in [d]$, all $a,b \in \mathbb{R}$, and value functions $v_1, v_2$, we have:
    \begin{align*}
        \phi_j(av_1 + bv_2) = a\phi_j(v_1) + b\phi_j(v_2),
    \end{align*}
    where $\phi_j(v)$ makes explicit the dependence of attributions on value functions. (Indeed, $\phi_j$ only depends on $i$ through $v$.) 
    \item \emph{Sensitivity}. If $v(i, S \cup \{j\}) = v(i, S)$ for all $S \subseteq [d] \backslash \{j\}$, then $\phi_j(i) = 0$.
    \item \emph{Symmetry}. If $v(i, S \cup \{j\}) = v(i, S \cup \{k\})$ for all $S \subseteq [d] \backslash \{j, k\}$, then $\phi_j(i) = \phi_k(i)$.
\end{itemize}
It can be shown that Shapley values are the unique solution to the attribution problem that satisfies efficiency, linearity, sensitivity, and symmetry \citep{Shapley1953}. Computing exact Shapley values is NP-hard, although numerous efficient approximations have been proposed \citep{Sundararajan2019}. 

There is some ambiguity as to how one should calculate payoffs on a proper subset of features, since $f$ requires $d$-dimensional input. Let $R = [d] \backslash S$, such that we can rewrite any $\bm{x}_i$ as a pair of subvectors $(\bm{x}_i^S, \bm{x}_i^R)$. Then the payoff for feature subset $S$ takes the form of a conditional expectation, with $\bm{x}_i^S$ held fixed while $\bm{x}^R$ varies. Following \citet{taly2020}, I consider a general form for the value function, parametrized by a distribution $\mathcal{D}_R$:
\begin{equation} \label{eq:value}
    v_{\mathcal{D}_R}(i, S) = \mathop{\mathbb{E}}_{\bm{x}^R \sim \mathcal{D}_R}[f(\bm{x}_i^S, \bm{x}^R)].
\end{equation}
Popular options for $\mathcal{D}_R$ include the marginal $P(\bm{X}^R)$, as in \citet{lundberg2017}; the conditional $P(\bm{X}^R|\bm{x}_i^S)$, as in \citet{Aas2019}; and the interventional $P(\bm{X}^R|do(\bm{x}_i^S))$, as in \citet{heskes2020}. I revisit the distinction between these distributions in Sect.~\ref{sec:ref_distro}.

Counterfactual explanations, unlike Shapley values, do not produce feature weights. Instead, they identify nearby datapoints with alternative outcomes -- the eponymous \emph{counterfactuals} -- intended to explain predictions by highlighting some (preferably small) set of input  perturbations. The optimization problem can be expressed as 
\begin{equation} \label{eq:cf}
    \bm{x}^* = \argmin_{\tilde{\bm{x}} \in \text{CF}(\bm{x}_i)} ~d(\bm{x}_i, \tilde{\bm{x}}),  
\end{equation}
where $\text{CF}(\bm{x}_i)$ denotes a counterfactual space, where $f(\bm{x}_i) \neq f(\tilde{\bm{x}})$ (for classification) or $f(\bm{x}_i) = f(\tilde{\bm{x}}) + \delta$ (for regression), where $\delta$ is a hyperparameter and $d$ is some distance measure (unspecified for now). The method was originally introduced by \citet{Wachter2018}, who use generative adversarial networks (GANs) to solve Eq.~\ref{eq:cf}. Others have proposed a variety of alternatives designed to ensure that counterfactuals are coherent and/or actionable \citep{Ustun2019, poyiadz_face, russell_coherent, karimi2020survey}. 

\subsection{Bayesian Decision Theory}
Why do we use XAI tools in the first place? \citet{Watson2021b} list three reasons: (1) to audit for potential bias; (2) to validate performance, guarding against unexpected errors; and (3) to discover underlying mechanisms of the data generating process. Another use case, with elements of (1) and (3) yet distinctly analyzed in the algorithmic recourse setting, is (4) to recommend actions so as to alter predicted outcomes. In all four cases, the aim is ultimately to make some sort of \textit{decision} -- be it about whether to sue a firm, deploy an algorithm, or perform an experiment. These motivations may overlap at the edges but they represent distinct tasks based on different assumptions and requiring their own explanatory methodologies. This heterogeneity is largely ignored by current feature attribution and counterfactual approaches, which implicitly assume a sort of \emph{explanatory objectivism}. According to this view, the quality of an explanation is independent of its context.

Decades of research in philosophy and social science has made such objectivism untenable \citep{Miller_2019, Floridi2019, sep-pragmatism}. If the view has any adherents today, perhaps it is because the alternative is often misconstrued as \emph{relativism}, an anything-goes slippery slope that inevitably leads to what \citet{Feyerabend1975} (approvingly) refers to as “epistemological anarchism”. This, however, is a false dichotomy. I reject the polar extremes of objectivism and relativism in favor of \emph{pragmatism}, which holds neither that there exists some single ideal explanation nor that all explanations are equally valid, but rather that various explanations may be more or less appropriate depending on who requests them and why. 

To operationalize this insight, I rely on the formal apparatus of decision theory. Let $A$ and $H$ denote finite sets of actions and outcomes, respectively. An agent’s preferences over action-outcome pairs induce a partial ordering that we represent with a utility function $u: A \times H \rightarrow \mathbb{R}$. Let $p(\cdot)$ denote a credence function over outcomes such that $p(H)=1$ and $p(h|E)$ denotes the agent’s subjective degree of belief in some $h \in H$ given a body of evidence $E = \{e_1, \dots, e_n\}$. The expected utility of each action $a \in A$ is given by a weighted average over hypotheses: 
\begin{equation}\label{eq:decision}
    \mathop{\mathbb{E}}_{h \sim p}[u(a,h)|E] = \sum_j p(h_j|E) ~u(a,h_j).
\end{equation}
If $u$ satisfies the utility axioms (i.e., completeness, transitivity, continuity, and independence) and $p$ satisfies the probability axioms (i.e., non-negativity, unit measure, and $\sigma$-additivity), then we say that the agent is \emph{rational}, and it can be shown that she will tend to maximize expected utility \citep{VonNeumann1944}. That is, she will always choose some optimal action:
\begin{equation}\label{eq:max_util}
    a^* = \argmax_{a \in A}~ \mathop{\mathbb{E}}_{h \sim p}[u(a,h)|E]
\end{equation}
from a set of alternatives. 

\sloppy This expected utility framework is very general, and applies to explanation instances of all four types cited above. For instance, in an auditing example where a data subject believes her loan application was denied due to some sensitive attribute(s), we might have $A = \{\texttt{sue}, \neg \texttt{sue}\}$ and $H = \{\texttt{biased}, \neg\texttt{biased}\}$, with evidence provided by some XAI tool and fairness criteria \citep{barocas-hardt-narayanan}. In algorithmic recourse, we typically consider more diverse action sets. For example, a credit scoring case may include $A = \{\texttt{education}, \texttt{income}, \texttt{credit}\}$, where increasing any subset of those variables incurs some cost but may improve outcomes $H = \{\texttt{approve}, \neg\texttt{approve}\}$.

A key consideration, often ignored by authors in XAI, is the difference between model- and system-level explanations \citep{Watson_conceptual}. At the model level, for instance, it is typically appropriate to treat all features as mutually independent and observe outputs on real and/or synthetic inputs that vary in some systematic fashion. At the system level, by contrast, we cannot afford to ignore causal relationships between features, as an intervention on one variable may have downstream effects on others. In these cases, nature constrains the input space in accordance with certain structural dependencies. 

This distinction is formalized using tools from the causal modeling literature \citep{Pearl2000}. A structural causal model (SCM) is a tuple $\mathcal{M} = \langle \bm{U}, \bm{V}, \mathcal{F} \rangle$, where $\bm{U}$ is a set of exogenous variables, i.e. background conditions; $\bm{V}$ is a set of endogenous variables, i.e. observed states; and $\mathcal{F}$ is a set of structural equations, one for each $V_j \in \bm{V}$. These functions map causes to direct effects in a deterministic fashion, $f_j: pa(V_j) \rightarrow V_j$, where $pa(\cdot)$ denotes the ``parents'' of some endogenous variable(s). The union of $V_j$'s direct and indirect causes (parents, grandparents, etc.) is called its ``ancestors'', denoted $an(V_j)$. 
Pearl's $do$-calculus \citeyearpar[Ch. 3.4]{Pearl2000} provides a provably complete set of rules for identifying causal effects from observational data where possible \citep{shpitser2008}.

A deterministic SCM can be expanded to accommodate stochastic relationships by placing a joint probability distribution on background conditions, $p(\bm{U})$. By equating this with the credence function introduced above, we may encode an agent’s uncertainty with respect to a causal system, which has implications for expected utility if $H \in \bm{V}$. This decision theoretic approach to causality is explicitly endorsed by \citet{Dawid2012, Dawid2015, dawid2020decisiontheoretic}, who argues that treatment policies should be designed not by considering differences between potential outcomes, but instead by minimizing expected loss as computed over a set of observed data. A similar premise underlies recent work in causal reinforcement learning \citep{lee2018, Lee_Bareinboim_2019}, where the goal is to find optimal interventions by maximizing expected rewards over trials. 

\section{Rational Shapley Values}\label{sec:method}

My basic strategy is to allow user inputs to inform the generation of counterfactuals and estimation of Shapley values, thereby resulting in custom feature attributions that are more useful than either method on its own. Intuitively, this gives agents the tools to pose not just generic (i.e., objectivist) questions of the form “Why did model $f$ predict outcome $\hat{y}_i$ for input $\bm{x}_i$?”, but more targeted (i.e., pragmatic) questions, such as: (1) Why did $f$ predict $\hat{y}_i$ as opposed to $\tilde{y}$ for $\bm{x}_i$? (2) Why did $f$ predict $\hat{y}_i$ as opposed to $\tilde{y}$ for $\bm{x}_i$, given $\bm{x}^S_i$? (3) Why did $f$ predict $\hat{y}_i$ as opposed to $\tilde{y}$ for $\bm{x}_i$, given $\bm{x}^S_i$ and certain constraints on $\bm{X}^R$? Unlike counterfactuals, which may confuse users by generating a large number of synthetic datapoints with no clear takeaway message, the proposed method provides unambiguous weights for all variables of interest. Unlike feature attributions, which ignore pragmatic information by construction, the present approach is \emph{rational} in the decision theoretic sense. I will make these claims more precise and apparent below. First, however, I introduce the notion of a \emph{relevant subspace} to formalize the hierarchy of specification implied by the questions above.
\begin{defn}[Relevant subspace]\label{def:rel_subsp}
    Let $\mathcal{Y}_0, \mathcal{Y}_1$ be a partition of $\mathcal{Y}$ into baseline and contrastive outcomes, respectively, with $\hat{y}_i \in \mathcal{Y}_0$. Let $S \subseteq [d]$ denote a (potentially empty) conditioning set with complement $R = [d] \backslash S$. Let $\mathcal{D}_S$ be the distribution of possible values for the conditioning set. Let $\mathcal{D}_R$ be a distribution encoding probable variation in $\bm{X}^R$ as a function of $\bm{x}^S$. Define $\mathcal{Z} = \mathcal{X} \times \mathcal{Y}$. We say that counterfactual sample $\tilde{\bm{z}} = (\tilde{\bm{x}}^S, \tilde{\bm{x}}^R, \tilde{y} = f(\tilde{\bm{x}}^S, \tilde{\bm{x}}^R))$ is \emph{relevant} for the inquiring agent if and only if: (i) $\tilde{y} \in \mathcal{Y}_1$; (ii) $\tilde{\bm{x}}^S \sim \mathcal{D}_S$; and (iii) $\tilde{\bm{x}}^R \sim \mathcal{D}_R$. Any space $\tilde{\mathcal{Z}} \subseteq \mathcal{Z}$ that meets these criteria is a \emph{relevant subspace}.
\end{defn}
The issue with popular Shapley value approximators (e.g., SHAP \citep{lundberg2017}) is that baselines are fixed by the data centroid. Attributions are only designed to account for the discrepancy between $f(\bm{x}_i)$ and the empirical mean $\phi_0(i) = n^{-1} \sum_{i=1}^n f(\bm{x}_i)$. Moreover, it is common to use the marginal $P(\bm{X}^R)$ as the reference distribution $\mathcal{D}_R$, thereby breaking associations between the features in $S$ and $R$, resulting in a value function that implicitly compares each $x_{ij}$ to the mean of the corresponding variable $X_j$. Yet these values might not fall within the relevant subspace for a given agent, who may, for instance, want to know why she received an average credit score instead of a better-than-average one. Or perhaps she wants to compare her predictions only to those of others in a similar age range and income bracket, effectively setting attributions for those features to zero. 

Counterfactuals naturally handle such cases, identifying real or synthetic datapoints that differ from the explanans in specific ways. However, there are several issues with the counterfactual approach. First, the cost function in Eq.~\ref{eq:cf} is almost always equated with some distance metric in practice, despite much discussion of more generic alternatives \citep{karimi2020survey, Ustun2019, Watson_lens}. This is an unsatisfying notion of costs, which can and should reflect the preferences of inquiring agents much more flexibly. For instance, relevant action sets may range over interventions exogenous to the feature space, where the notion of distance is often inapplicable.
Second, there is the question of how many counterfactuals to provide. Eq.~\ref{eq:cf} suggests that just a single sample would suffice; however, such a procedure overestimates the certainty with which such a datapoint is generated. Can we really be sure that counterfactual $\bm{x}^*$ is to be preferred to counterfactual $\bm{x}'$ just because the former is marginally closer in feature space to the input $\bm{x}_i$? Even if we are willing to accept a purely distance-based notion of cost, it is not at all obvious that very small distances should be taken seriously when there are so many potential sources of noise – in the measurements themselves, in the finite training data, and in whatever approximations were necessary to efficiently generate candidate counterfactuals. That is why some authors prefer to sample a large, diverse coalition of counterfactuals \citep{mothilal2020_dice}, so that users can survey the various paths through which a prediction may change. Yet this approach raises new problems of intelligibility. How should an agent make sense of a relevant subspace with dozens or hundreds of samples within a tolerable distance of $\bm{x}_i$? How should she prioritize among different counterfactuals or integrate this information into a high-level summary?

Feature attribution methods are well suited to this task. Shapley values provide an efficient and intuitive account of the information encoded in a large set of counterfactuals, regardless of sample size. I therefore propose a hybrid method in which data are sampled from a relevant subspace, and Shapley values computed only with respect to such samples. The crucial observation is that nothing in the definition of Shapley values (Eqs.~\ref{eq:additive}, \ref{eq:shapley}, \ref{eq:value}) precludes us from shifting the reference distribution from an entire observational dataset to a particular region with certain desirable properties -- i.e., a relevant subspace. 

\subsection{Selecting the Reference Distribution}\label{sec:ref_distro}
I take it more or less for granted that agents requesting explanations for algorithmic predictions have some contrastive outcome $\mathcal{Y}_1$ in mind. They are also likely to have some intuition about the conditioning set $S$, which will presumably include immutable features and/or any variables that the agent does not want to enter into the explanation if it can be avoided. The precise distribution for these features may not be available \emph{a priori}, but conditioning on the observed $\bm{x}_i^S$ is probably a good start when the data allow for it. The trickiest component to evaluate then is the reference distribution $\mathcal{D}_R$, which has implications both for how counterfactuals are computed (since this helps define the relevant subspace) and for how Shapley values are estimated (since this helps define the value function). 

I consider three possibilities: (1) marginal: $P(\bm{X}^R)$; (2) conditional: $P(\bm{X}^R|\bm{x}^S_i)$; and (3) interventional: $P(\bm{X}^R|do(\bm{x}^S_i))$. The first point to observe is that the three distributions are not equivalent. (1) comes apart from the other two whenever statistical dependencies are present, i.e. $\bm{X}^R \dep \bm{X}^S$. Options (2) and (3) come apart whenever causal dependencies are present, i.e. if $an(\bm{X}^R) \cap \bm{X}^S \neq \{\emptyset\}$. I want to be clear that the differences between these distributions are in no sense normative. None is generally ``better'' than the others, or more useful, or more broadly applicable. Rather, each is the right answer to a different question. What follows is a brief discussion of the advantages and disadvantages associated with each value of $\mathcal{D}_R$, as well some heuristic guidelines on when to use which.


\emph{Marginal}. The marginal distribution is preferable when the goal is a model-level explanation. This may be the case in certain instances of auditing or validation, where the analyst seeks merely to discover what the model has learned without any further restrictions. In this case, we may not care too much about impossible data perturbations (e.g., teenage grandmothers) since we simply want to recreate a decision boundary in model space. The extent to which this model space matches reality is another question altogether, presumably one that data scientists should take seriously, both before and throughout deployment. Once the model is in the wild, however, it may be of independent interest to users and regulators how it performs on a wide range of hypothetical cases, including those off the true data manifold.

\emph{Conditional}. The conditional distribution is better suited to explanations at the system level, where impossible data permutations could lead to issues. This may be the case, for instance, in certain instances of validation, where we do not wish to punish a model for failing to extrapolate to data points far from its training set. Such methods are especially valuable when joint distributions may be estimated with reasonable accuracy and causal relationships are unknown or potentially inapplicable (e.g., in image classification tasks). Estimating joint densities is a hard problem in general, especially when data are high-dimensional and/or of mixed variable types. 

\emph{Interventional}. The interventional distribution is optimal when seeking explanations at the system level for causal data generating processes. With access to the underlying SCM, there can be no more accurate estimator than that defined by (3). These methods are required when seeking to use XAI for discovery and/or planning, as both sorts of actions invariably rely on real-world mechanisms that cannot be approximated by either a purely marginal approach (which ignores all dependencies) or a conditional one (which fails to distinguish between correlation and causation). Of course, complete causal information is almost never available, which means that in practice analysts must make strong assumptions and/or make do with workable approximations.

\sloppy I employ marginal, conditional, and interventional value functions in all experiments below (see Sect.~\ref{sec:exp}). To compute interventional Shapley values, I rely on partial orderings, as these are often the most readily available to analysts at test time. A partial ordering of features implies a chain graph in which links are composed of all variables in a single causal group. For instance, users may not know the complete graph structure of the set $\{\texttt{age}, \texttt{sex}, \texttt{income}, \texttt{savings}, \texttt{credit}\}$ but they may be more confident about the partial ordering $\{\texttt{age}, \texttt{sex}\} \rightarrow \{\texttt{income}, \texttt{savings}\} \rightarrow \{\texttt{credit}\}$. \citet{Lauritzen2002} demonstrate how to factorize the probability distribution for a DAG of chain components. \citet{heskes2020} combine this procedure with a few steps of $do$-calculus to compute interventional Shapley values from observational data.

\subsection{A New Axiom}
Say two agents are identical along all recorded variables for some credit scoring task, and both receive lower scores than they had hoped. Counterfactuals identify two minimal perturbations sufficient to improve predictions: (1) increase education by one unit; or (2) increase savings by one unit. Let us stipulate that these actions are equivalent in terms of time and money. However, they differ in another important respect -- the first agent wants to go back to school and the second wants to save more each month. In this case, \emph{preferences alone} determine which of two explanations is optimal. 

We may devise a similar example to show the impact of differing credences. Say, for instance, that two labs are planning expensive gene knockout experiments to test the predictions of a gene regulatory network inference algorithm. Their utilities may be identical -- the goal for both groups is to maximize true discoveries while minimizing false positives -- but they are working with different sets of evidence after preliminary, as yet unpublished results indicated to the first lab that one set of pathways is a probable dead end. In this case, \emph{beliefs alone} determine which of two explanations is optimal.

What these examples demonstrate is that utilities and credences matter when attempting to explain model predictions. With default software like SHAP, feature attributions for these credit loan applicants or research labs would not generally differ -- same inputs, same outputs. However, by explicitly incorporating pragmatic information, we can secure custom explanations at no cost to the axiomatic guarantees of Shapley values. In fact, we can extend the current desiderata.

Observe that for fixed value function $v$, model $f$, and input $\bm{x}_i$, feature attributions $\bm{\phi}= \{\phi_j\}_{j=1}^d$ vary only as a function of contrastive baseline $\tilde{\bm{z}}$. I make this dependence explicit moving forward, writing $\bm{\phi} (\tilde{\bm{z}})$. Define a reward function for the inquiring agent as the expected utility of the utility-maximizing action $a^*$, conditional on a given set of feature attributions:
\begin{equation}
    r(\tilde{\bm{z}}) = \mathop{\mathbb{E}}_{h \sim p}[u(a^*, h) | \bm{\phi} (\tilde{\bm{z}})].
\end{equation}
The axiomatic constraint that characterizes pragmatic feature attributions can now be stated as follows. For the proof of Thm.~\ref{thm:rational}, see Appx.~\ref{sec:appx}.
\begin{defn}[Rationality]\label{def:rational}
    Let $r(\cdot)$ be the reward function for a rational agent, i.e. one who behaves in accordance with the utility and probability axioms. Let $\bm{\phi}(\tilde{\bm{z}})$ be a feature attribution vector computed with respect to subspace $\tilde{\mathcal{Z}}$. Then $\bm{\phi}(\tilde{\bm{z}})$ is \emph{rational} if and only if, for any alternative subspace $\mathcal{Z}'$, $\mathbb{E}_{\tilde{\bm{z}} \sim \tilde{\bm{Z}}} [r(\tilde{\bm{z}})] \geq \mathbb{E}_{\bm{z}' \sim \mathcal{Z}'}[r(\bm{z}')]$. In other words, feature attributions are rational to the extent that they tend to maximize expected rewards for the inquiring agent.
\end{defn}
\begin{theorem}\label{thm:rational}
    When the relevant subspace is nonempty, rational Shapley values are the unique additive feature attribution method that satisfies efficiency, linearity, sensitivity, symmetry, and rationality. 
\end{theorem}

\subsection{Rational Shapley Value Algorithm}\label{sec:algo}
\begin{algorithm}[H]
    \caption{Rational Shapley Values}
    \KwIn{Input datapoint $\bm{x}_i$, set of candidate subspaces $\{\langle \mathcal{D}_S, \mathcal{D}_R, \mathcal{Y}_1 \rangle_k\}_{k=1}^m$, utility function $u$, credence function $p$}
    \KwOut{$\bm{\phi}(\tilde{\bm{z}}^*)$}
    Compute classical Shapley values $\bm{\phi}(\bm{z})$ using mean $\phi_0 = n^{-1} \sum_{i=1}^n f(\bm{x}_i)$ and reference $\mathcal{D}_R$\\
    Initialize $r(\bm{z}) \leftarrow \sum_j p(h_j|\bm{\phi}(\bm{z})) ~u(a^*,h_j)$\\
    \For{$k = \{1, \dots, m\}$}{
    Draw data $\tilde{\bm{z}}_k \sim \langle \mathcal{D}_S, \mathcal{D}_R, \mathcal{Y}_1 \rangle_k$\\
    \eIf{$\tilde{\bm{z}}_k = \{\emptyset\}$}{
    \tt{Fail}}{
    Compute $\bm{\phi}(\tilde{\bm{z}}_k)$\\
    Record $r(\tilde{\bm{z}}_k) \leftarrow \sum_j p(h_j|\bm{\phi}(\tilde{\bm{z}})) ~u(a^*,h_j)$}}
    $\tilde{\bm{z}}^* \gets \argmax_{\tilde{\bm{z}}} r(\tilde{\bm{z}})$
\label{alg:rat_shap}
\end{algorithm}

I reframe the objective function of Eq.~\ref{eq:cf} with a slight twist. Our goal is not to minimize the cost or distance between two individual datapoints, but rather to compute a subspace that maximizes rewards. That is:
\begin{equation}\label{eq:obj}
    \mathcal{Z}^* = \argmax_{\tilde{\mathcal{Z}} \subseteq \mathcal{Z}}~ \mathop{\mathbb{E}}_{\tilde{\bm{z}} \sim \tilde{\mathcal{Z}}}[r(\tilde{\bm{z}})].
\end{equation}
This differs from (and improves upon) Eq.~\ref{eq:cf} in several respects. First, it results not in a single point but in a region of space, which is stabler and more informative. Second, it explicitly incorporates preferences and credences via the functions that define $r(\cdot)$. Finally, since rewards are conditioned upon a feature attribution vector, this target combines elements of both counterfactuals and feature attributions, which is preferable to either alternative alone for all the reasons argued above.

There are many ways one could go about computing this subspace in practice, and no single method is optimal in general. See \citet{karimi2020survey} for a recent review. In the following experiments, I follow \citet{Wexler2020} in relying on observational samples. These subspaces have the advantage of being fast to find and guaranteed to lie on the true data manifold (up to potential sampling or measurement errors). They avoid the assumptions and approximations of more convoluted optimization techniques. The main disadvantage of this method is that it may provide low coverage for regions of the feature space where data are undersampled. However, this poses issues for other approaches as well, as no method can confidently draw realistic points from low-density regions of the feature space. 

The basic strategy is schematized in pseudocode (see Alg.~\ref{alg:rat_shap}). The user postulates a set of candidate subspaces, computing a sequence of rational Shapley vectors and comparing relative rewards. Evaluating $r(\tilde{\bm{z}})$ with precision requires explicit utilities and (conditional) credences, which may not be generally available. This can be done informally by comparing feature attribution vectors on an ordinal basis, such that rewards are ranked rather than directly quantified. This procedure is equivalent to the cardinal alternative under reasonable assumptions about agentive rationality, as noted above. 
The algorithm is deliberately silent on how to select candidate subspaces, as this is irreducibly context-dependent. Credence updates may be calculated in closed form under some parametric assumptions, or else via Monte Carlo. The examples below illustrate how the method works in practice.


\section{Experiments}\label{sec:exp}
In this section, I describe results from a number of experiments on benchmark datasets. I compare performance against baselines using marginal, conditional, and interventional value functions, respectively labelled MSV, CSV, and ISV. Code for reproducing all results and figures can be found at \url{https://github.com/dswatson/rational_shapley}.

\subsection{Auditing: COMPAS}
The COMPAS algorithm is a statistical model used to assign risk scores to defendants awaiting trial. High risk scores are associated with elevated probability of recidivism, which judges in nine US states use to help decide whether to let defendants out on bail while awaiting trial. A 2016 report by ProPublica alleged that the COMPAS algorithm was racially biased against African Americans \citep{Angwin2016}; subsequent analysis by independent researchers has not always corroborated those findings \citep{Fisher2019, Rudin2020Age}.

Though the creators of COMPAS have published technical reports defending their model, they have made neither their training data nor code publicly available, so attempts to recreate the algorithm and audit for racial bias are typically based on data gathered by ProPublica on some 12,000 individuals arrested in Broward County, Florida between 2013 and 2014. Using a random forest \citep{Breiman2001}, I regress violent risk scores on the following features: \texttt{age}, number of \texttt{priors}, and whether the present charge is a \texttt{felony} (all deemed “admissible”); as well as \texttt{race} and \texttt{sex} (deemed “inadmissible”). For the purposes of this experiment, I take the trained model as the true target. 

MSV is computed via Monte Carlo \citep{Strumbelj2014}, with 2000 simulations per attribution. CSV is computed using the TreeSHAP method \citep{Lundberg2020}, since random forests are composed of individual regression trees. For ISV, I presume a simple partial ordering in which demographic variables (\texttt{age}, \texttt{sex}, and \texttt{race}) are root nodes, upon which all other predictors depend. The mean predicted response for all defendants in the dataset is $\phi_0 = 3.69$.

\begin{figure}[h]
\begin{center}
\includegraphics[width=.6\textwidth]{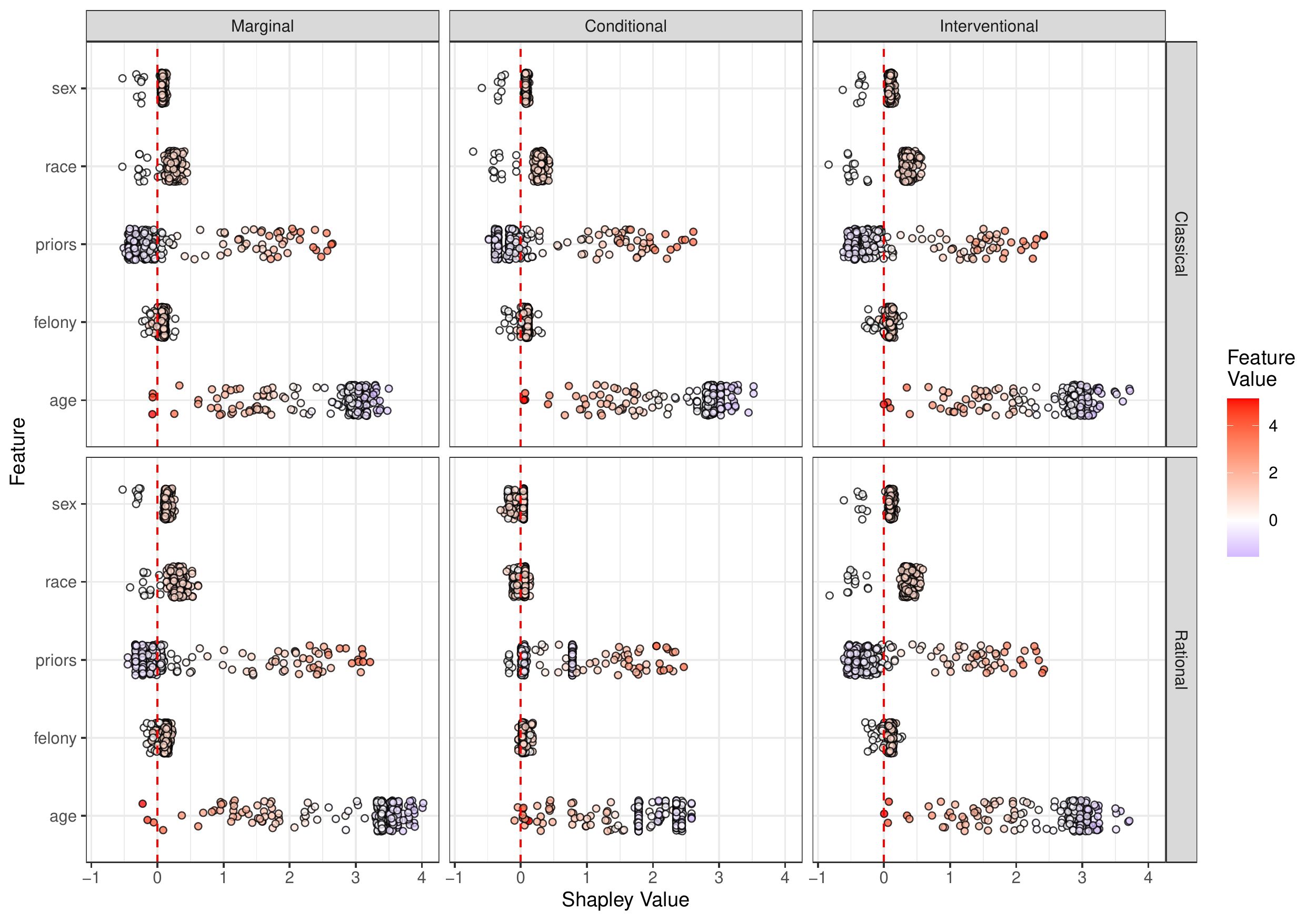}
\end{center}
\caption{Classical and rational Shapley values for the Broward County COMPAS dataset, computed with marginal, conditional, and interventional reference distributions. Continuous feature values are $z$-scored for visualization.}
\label{fig:compas_shap}
\end{figure}

I focus specifically on high-risk defendants, namely the 264 subjects with risk scores at or above the $95^{\text{th}}$ percentile according to the COMPAS approximator. Some 94\% of these defendants are African American, compared to just 60\% in the general dataset, which immediately raises questions about racial bias. MSV, CSV, and ISV all look fairly similar in this case (see Fig.~\ref{fig:compas_shap}). Age and priors generally receive the highest feature attributions in this group, consistent with the findings of other researchers. Race also receives nonzero attribution throughout. Notably, race receives higher attribution under the ISV reference than MSV, since MSV can only detect direct effects, and race is presumed to have both direct and indirect effects given the partial ordering. In this example, I presume that auditors are working at the model level -- i.e., evaluating the patterns learned by the algorithm, rightly or wrongly -- in which case the marginal distribution is best suited to their purpose. 

To compute rational Shapley values, I consider a relevant subspace of defendants in the same age range (at most 33 years old) with at least one prior, yet who fell in the bottom half of defendants by predicted risk score. All other features are allowed to vary unconstrained. 356 subjects in the data meet these criteria, and therefore constitute the reference sample for this experiment. This is too large a group of counterfactuals to “explain” anything on their own, even if they were ranked by some measure (e.g., distance from the target input). What the task requires is a summary of differences across features, a job for which Shapley values are uniquely well-suited. The racial breakdown of defendants in the relevant subspace is markedly different from our high-risk group -- just 48\% are African American -- further reinforcing concerns about potential bias in the COMPAS data. The resulting Shapley values look fairly similar to their classical counterparts, although priors and age appear to dominate the explanations under conditional and interventional reference distributions. Using the marginal reference, however -- which, as noted above, is most relevant for auditing scenarios -- we find greater evidence of racial bias using rational Shapley values. The racial attribution gap between Black and white defendants is significantly greater in the relevant subspace than it is across the complete dataset $(t = 3.874, p < 0.001)$.


Consider the case of a young Black defendant deciding whether to file suit against the makers of COMPAS. Despite having just a single prior offense, this 21-year-old was placed in the highest risk group according to the algorithm. A utility matrix for this individual is given in Table~\ref{tbl:1}. Assume that the defendant in question assigns a uniform prior over $H$ to begin with. We say that $h_1$ is corroborated to the extent that race receives greater positive attribution for African Americans than it does for defendants of other races. Then although both classical and rational Shapley values recommend the same action $(a_1)$, the latter does so with higher expected reward, since $p(h_1 | \bm{\phi}(\tilde{\bm{z}})) > p(h_1 | \bm{\phi}(\bm{z}))$. Rational Shapley values are therefore preferable in this case, as theory suggests.

\begin{table}
\caption{Utility matrix for a 21-year-old African American defendant with a single prior, predicted to be high risk and deciding whether to sue the makers of COMPAS.}
    \centering
    \begin{tabular}{|c|c|c|}
    \hline
         & $h_1: \texttt{biased}$ & $h_2: \neg \texttt{biased}$ \\ \hline
        $a_1: \texttt{sue}$ & 5 & –1 \\ \hline
        $a_2: \neg \texttt{sue}$ & 0 & 0 \\ \hline
    \end{tabular}
\label{tbl:1}
\end{table}

\subsection{Discovery: Medical Diagnosis}
As an example of XAI for discovery, I consider the diabetes dataset originally described by \citep{Efron2004}. The data consist of 442 patients and 10 predictors -- \texttt{age}, \texttt{sex}, body mass index (\texttt{BMI}), blood pressure (\texttt{MAP}), and six blood serum measurements, including three cholesterol-related variables (\texttt{LDL}, \texttt{HDL}, and \texttt{TC}). The goal is to predict disease progression after one year, for which I use an elastic net regression \citep{Zou2005}. Note that exact Shapley values can be analytically calculated for linear models under the assumption of feature independence. However, exploratory data analysis reveals strong correlations between some features in the data (especially blood serum measurements), and causal dependencies between features such as BMI and MAP are well established. In such a case, the analytic formulae for local explanation in linear models are inapplicable. For ISV, I assume the partial ordering $\{\texttt{age}, \texttt{sex}, \texttt{bmi}\} \rightarrow \{\texttt{map}, \texttt{ldl}, \texttt{hdl}, \texttt{tc}\} \rightarrow \{\texttt{tch}, \texttt{ltg}, \texttt{glt}\}$. This treats age, sex, and BMI as root nodes; MAP and cholesterol as causally intermediate; and remaining blood serum measurements are downstream. Such a DAG admittedly oversimplifies several complex biochemical processes but is broadly consistent with known structural relationships. 

\begin{figure}[t]
\begin{center}
\includegraphics[width=.6\textwidth]{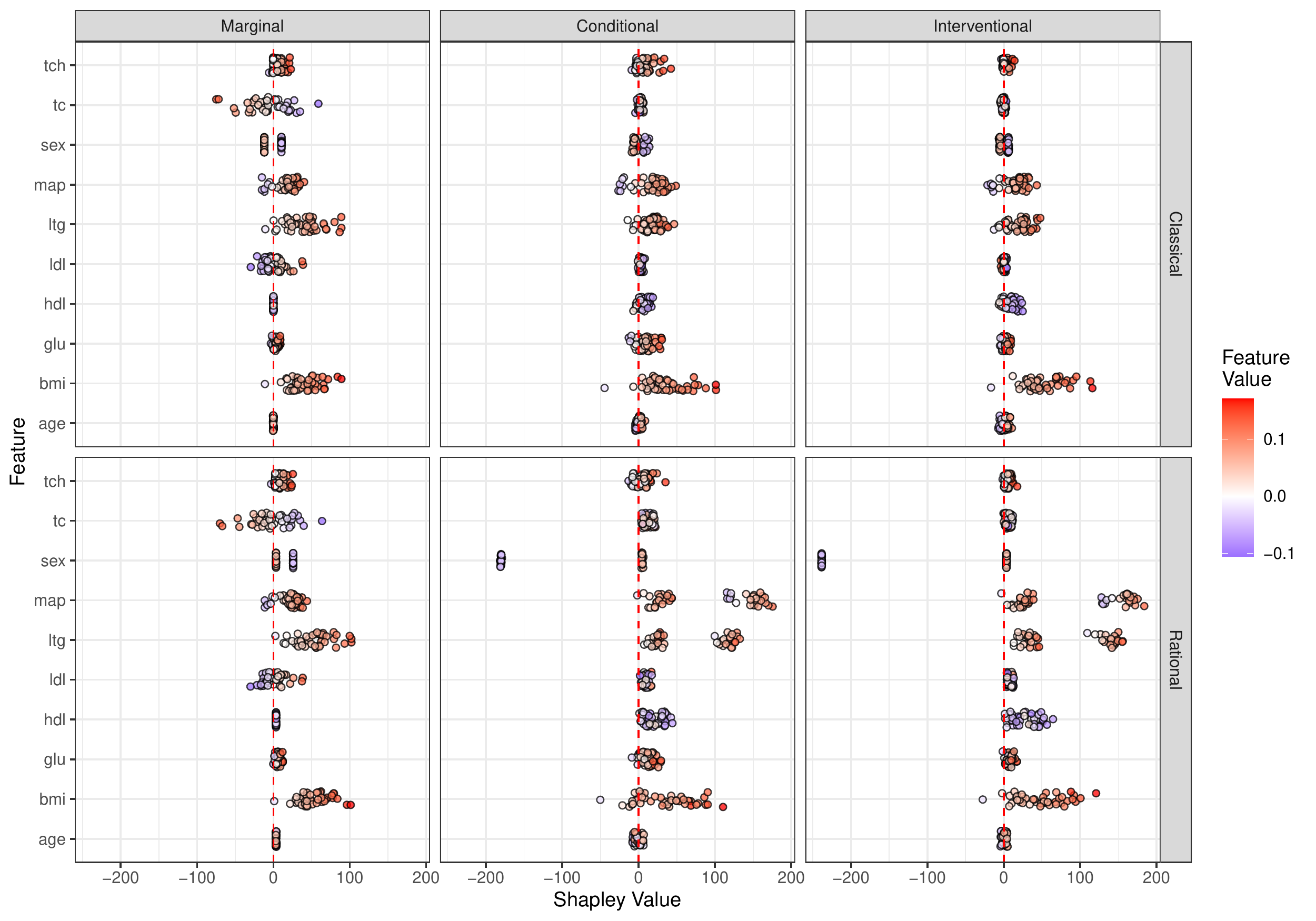}
\end{center}
\caption{Classical and rational Shapley values for the diabetes dataset, computed with marginal, conditional, and interventional reference distributions. Continuous feature values are $z$-scored for visualization.}
\label{fig:medical}
\end{figure}

I focus on patients with especially poor prognoses, as these are typically the subjects of greatest concern to clinicians. Specifically, I examine the top decile of patients by disease activity $(\hat{y} \geq 265)$. Classical Shapley values for these patients are visualized under all three value functions in Fig.~\ref{fig:medical}. CSV and ISV are fairly similar here, although both are quite different from the marginal alternative. For instance, \texttt{TC} receives relatively large attributions under the marginal reference distribution, but nearly none under conditional or interventional distributions, owing to its strong collinearity with other cholesterol measures. 

For rational Shapley analysis, I consider a subspace consisting of older men (i.e., those in the top quartile by age), who are generally at higher risk than other groups in this dataset. Since age and sex are (more or less) unactionable, a rational agent will likely want prognostic explanations that place little or no weight on such features. We find in Fig.~\ref{fig:medical} that BMI and MAP receive large rational attributions on average, reflecting their greater variance in the relevant subspace. Conditional and interventional reference distributions are especially valuable here, as the analytical goal almost certainly requires a system-level approach and causal dependencies cannot therefore be ignored. Alternative structural assumptions may lead to somewhat different attributions, but focusing on a relevant subspace has effectively driven the Shapley values for unactionable variables like sex and age toward zero. The result is that explanations are provided primarily in terms of lifestyle variables like BMI and MAP, which agents can work to minimize in order to improve their disease prognosis.

\begin{table}
\caption{Complete feature vector for both Bert and Ernie in the diabetes dataset.}
    \centering
    \begin{tabular}{|c|c|c|c|c|c|c|c|c|c|}
    \hline
        \texttt{sex} & \texttt{age} & \texttt{bmi} & \texttt{map} & \texttt{tc} & \texttt{ldl} & \texttt{hdl} & \texttt{tch} & \texttt{ltg} & \texttt{glu} \\ \hline
        1 & –0.064 & 0.096 & 0.105 & –0.003 & –0.005 & –0.007 & -0.003 & 0.023 & 0.073 \\ \hline
    \end{tabular}
\label{tbl:2}
\end{table}

Consider the case of two high-risk patients -- call them Bert and Ernie -- who share identical feature vectors (see Table~\ref{tbl:2}; note that the negative value of \texttt{age} is a quirk of the scaling procedure.) Despite their striking similarities, however, Bert and Ernie differ in their culinary tastes. Bert was raised on carbohydrate-rich foods, and cannot imagine living without bread, pasta, and potatoes; Ernie, by contrast, is a voracious carnivore who eats bacon for breakfast, cold cuts for lunch, and steak for dinner. Their respective utility matrices are given by Table~\ref{tbl:3}, where we assume for the sake of this example that only two dietary interventions are under consideration: a low-carb diet and a low-fat diet. Evidence suggests that the latter is slightly more effective for weight loss, while the former has the added benefit of reducing blood pressure \citep{Brinkworth2009}. Classical Shapley values, which place the greatest weight on BMI, would tend to favour $a_1$ regardless of agentive preferences. However, rational Shapley values can distinguish between optimal explanations based entirely on utilities, since changes to either BMI or MAP are sufficient to bring about the desired outcome, albeit at differing costs to Bert and Ernie. Thus we find the following two feature attribution vectors (see Table~\ref{tbl:4}), computed using the conditional reference distribution on inputs that differ only with respect to utilities $u$.

\begin{table}
\caption{Utility matrices for Bert (left) and Ernie (right). Bert would like to reduce his risk of diabetes without lowering his carb intake; Ernie wants the same outcome without lowering his fat intake.}
\begin{subtable}{.5\linewidth}
    \centering
    \begin{tabular}{|c|c|c|}
    \hline
        \textbf{Bert} & $h_1: \hat{y} < 265$ & $h_2: \hat{y} \geq 265$ \\ \hline
        $a_1$: low-fat & 5 & –1 \\ \hline
        $a_2$: low-carb & 1 & –6 \\ \hline
    \end{tabular}
\end{subtable}%
\begin{subtable}{.5\linewidth}
    \begin{tabular}{|c|c|c|}
    \hline
        \textbf{Ernie} & $h_1: \hat{y} < 265$ & $h_2: \hat{y} \geq 265$ \\ \hline
        $a_1$: low-fat & 1 & –6 \\ \hline
        $a_2$: low-carb & 5 & –1 \\ \hline
    \end{tabular}
\end{subtable}%
\label{tbl:3}
\end{table}

\begin{table}
\caption{Rational Shapley vectors for Bert (top) and Ernie (bottom), computed from the same input vector (see Table \ref{tbl:2}) but with respect to different relevant subspaces.}
    \centering
    \begin{tabular}{|c|c|c|c|c|c|c|c|c|c|}
    \hline
        \texttt{sex} & \texttt{age} & \texttt{bmi} & \texttt{map} & \texttt{tc} & \texttt{ldl} & \texttt{hdl} & \texttt{tch} & \texttt{ltg} & \texttt{glu} \\ \hline
        –3.789 & –4.867 & 94.922 & 29.241 & 1.763 & 0.849 & 4.053 & 6.566 & 13.019 & 4.324 \\ \hline
        –1.291 & 0.114 & 42.604 & 66.942 & 2.187 & 1.129 & –0.167 & –3.532 & 11.136 & 9.903 \\ \hline
    \end{tabular}
\label{tbl:4}
\end{table}

Observe that, among the many differences between these two vectors, they flip the relative importance of BMI and MAP, suggesting two alternative paths toward reducing disease risk. A counterfactual approach with a well-designed cost function could in principle also identify different explanations for these two agents, but it could not additionally provide complete feature attributions summarising the relative impact of other predictors. Such a synthesis is unique to the rational Shapley approach. 

\subsection{Recourse: Credit Scoring}
As a final example, I examine the recourse setting, where the goal is to advise an agent what interventions are necessary and/or sufficient to secure a desired prediction. As \citet{karimi2021} have shown, this task requires causal information. More specifically, they prove that noncausal recommendations, such as those provided by \citet{Ustun2019}, guarantee recourse if and only if the treatment variables have no descendants in the underlying causal graph. For this experiment, I use the German credit dataset from the UCI Machine Learning Repository \citep{Dua2017}. To simplify the presentation, I restrict focus to just seven of the most informative features: age $A$, gender $G$, marital status $M$, job $J$, savings $S$, loan amount $L$, and duration $D$. The outcome $Y$ is binary (loan approved/denied), with a base rate of 70\%. I train a support vector machine (SVM) \citep{Cortes1995} to predict $Y$ using a radial basis kernel.

\begin{figure}[t]
\begin{center}
\includegraphics[width=.6\textwidth]{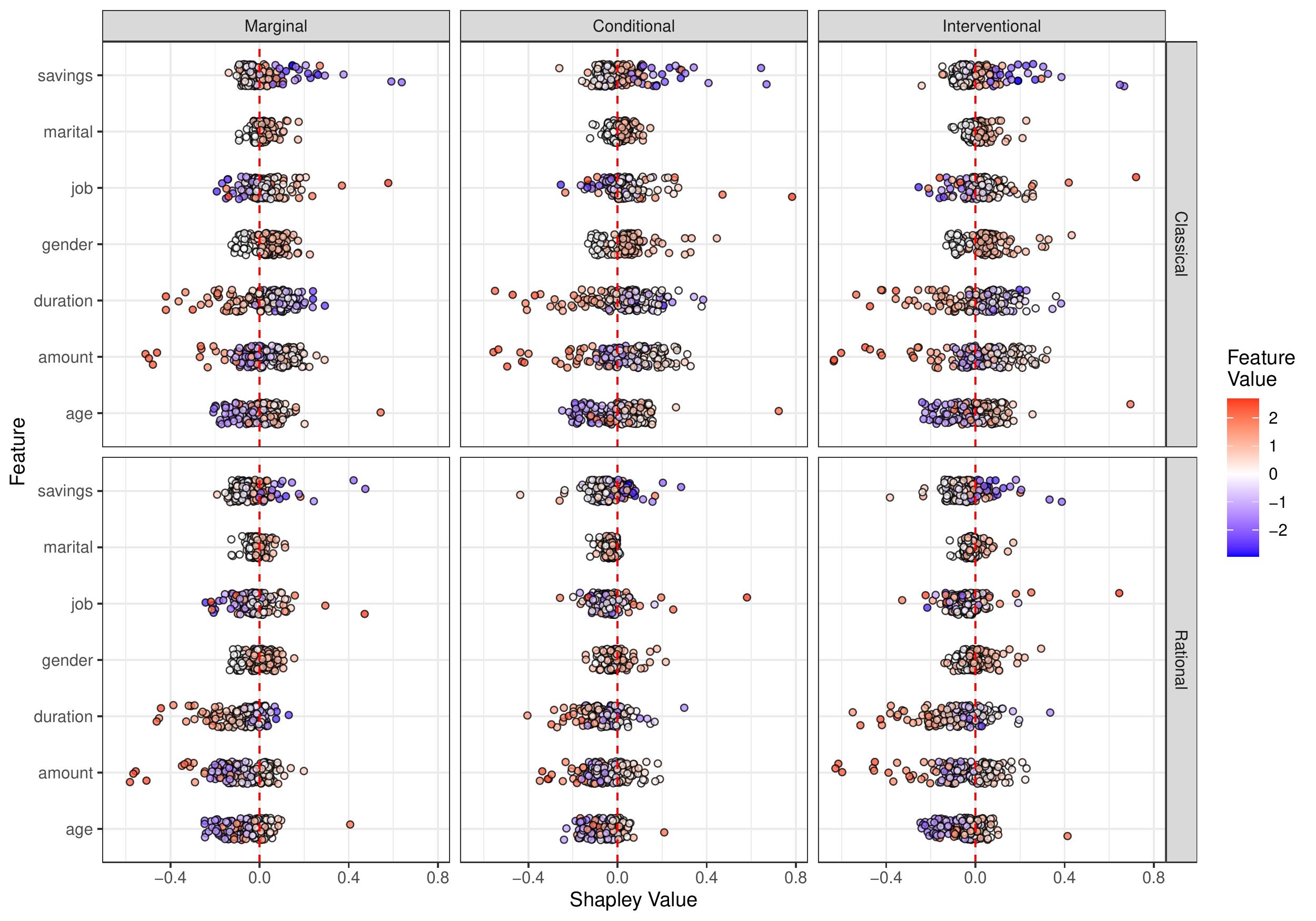}
\end{center}
\caption{Classical and rational Shapley values for the German credit dataset, computed with marginal, conditional, and interventional reference distributions. Continuous feature values are $z$-scored for visualization.}
\label{fig:credit}
\end{figure}

I assume a simple partial ordering in which demographic variables are causally antecedent to both financial predictors and loan application details: $\{A, G, M\} \rightarrow \{L, D, J, S\}$. The UCI website notes that false positives are more costly for banks than false negatives, and therefore recommends penalizing the former at five times the rate of the latter. Post-processing the SVM, an optimal decision threshold is obtained at 0.74, and I therefore label all and only those points with predicted probabilities at or above this value as 1. The resulting model is about 69\% accurate on the true positives and 73\% accurate on the true negatives. I focus on the 189 borderline applicants with predicted success probabilities on the interval $[0.7, 0.74)$. These are subjects who could plausibly benefit from even a slight improvement in their model predictions. Classical Shapley values are unable to help such applicants, since the mean response in the full dataset is $\phi_0 = 0.7$, i.e. the lower bound of the interval. (Irrational) feature attributions for these subjects are therefore guaranteed to sum to some value on $[0, 0.04)$, explaining why they did slightly better than average instead of why they were ultimately unsuccessful.

The most important component of the relevant subspace is the contrastive outcome $\mathcal{Y}_1$, as the borderline applicants appear fairly representative along all predictors of interest. I therefore stratify by predicted response, zooming in on the top quartile of applicants $(\hat{y} \geq 0.76)$. Bivariate analyses suggest that such applicants are for the most part financially secure and request small loans of relatively brief duration. The classical and rational Shapley vectors look fairly similar in this experiment, except for the crucial difference in offset. (See Fig.~\ref{fig:credit}; note that the response variable is transformed to the logit scale for easier visualization.) 

Consider the case of a loan applicant, call her Ruth, whose predicted success probability is 0.73 -- just under the decision boundary. Her input feature vector can be found in Table~\ref{tbl:5}, where values are reported on the original (pre-transformation) scale. She is looking to buy a new home and believes her high-skilled job and decent savings (both well above their respective median values in the dataset) should make her a strong applicant. She is willing to make some changes if necessary -- specifically, to increase her savings, decrease the size of her requested loan, or pay it back sooner -- but in any case she would like to act quickly, since she fears the house she wants to buy will not be on the market for long. Ruth’s utility matrix is given in Table~\ref{tbl:6}. We find here that she prefers changing the terms of her application to increasing her savings in general, as this is likely the faster route to bank approval. She is indifferent between changing the loan amount or duration. 
As noted above, computing Shapley values from the entire dataset is pointless in this case, as Ruth’s predicted outcome is already above the base rate. Counterfactuals could potentially aid her search for an explanation, but only by either providing an artificially narrow set of pathways to approval or overwhelming her with an unnecessarily large reference class of successful applicants. 

\begin{table}
\caption{Complete feature vector for Ruth in the German credit dataset.}
    \centering
    \begin{tabular}{|c|c|c|c|c|c|c|}
    \hline
        \texttt{gender} & \texttt{marital} & \texttt{age} & \texttt{amount} & \texttt{duration} & \texttt{savings} & \texttt{job} \\ \hline
        0 & 0 & 26 & 3181 & 26 & 290 & 2 \\ \hline
    \end{tabular}
\label{tbl:5}
\end{table}

\begin{table}
\caption{Utility matrix for Ruth, whose demographic and financial information place her just below the loan approval threshold for a hypothetical bank.}
    \centering
    \begin{tabular}{|c|c|c|}
    \hline
         & $h_1:\hat{y} \geq 0.74$ & $h_2: \hat{y} < 0.74$ \\ \hline
        $a_1$: Increase $S$ & 1 & –2 \\ \hline
        $a_2$: Decrease $L$ & 2 & –1 \\ \hline
        $a_3$: Decrease $D$ & 2 & –1 \\ \hline
    \end{tabular}
\label{tbl:6}
\end{table}

Rational Shapley values, computed from a relevant subspace of successful loan applicants who share Ruth’s demographic characteristics, provide a summary of feature attributions within this reference class (see Table~\ref{tbl:7}; note that calculations are performed in logit space.) Interestingly, we find that her requested loan amount, though above average for the relevant subspace, did not hurt Ruth’s application. On the contrary, it appears to boost her chances somewhat, a result we might not expect from surveying the counterfactuals directly. The loan duration, however -- about double the average for the relevant subspace -- brought her success probability down considerably. This is a highly actionable piece of information, insomuch as it guides Ruth toward algorithmic recourse that can push her application over the decision boundary. Of course, if the change is too onerous for Ruth, then she is free to resample from another subspace of successful applicants with longer loan duration on average. The procedure may continue like this indefinitely, with agents testing out new hypotheses in an iterative, exploratory fashion, updating their rewards accordingly. 

\begin{table}
\caption{Rational Shapley values for Ruth in the German credit dataset.}
    \centering
    \begin{tabular}{|c|c|c|c|c|c|c|}
    \hline
        \texttt{gender} & \texttt{marital} & \texttt{age} & \texttt{amount} & \texttt{duration} & \texttt{savings} & \texttt{job} \\ \hline
        0 & 0 & –0.168 & 0.061 & –0.115 & 0.013 & 0.128 \\ \hline
    \end{tabular}
\label{tbl:7}
\end{table}

\section{Discussion}\label{sec:discussion}
The skeptical reader may plausibly object that the examples above are fairly neat and straightforward, with their small utility matrices of well-defined user preferences. Reality, of course, is far messier. Complete sets of actions and outcomes may not be known upfront, let alone utility and credence functions defined thereon. Potential interventions may be far more numerous than these experiments permit, and associated outcomes entirely uncertain. Can rational Shapley values scale to larger, more complicated instances of algorithmic explanation?

In a word, yes. These cases are merely illustrative, following on the back of theoretical results establishing the viability of a pragmatic synthesis between feature attributions and counterfactuals that works with various different reference distributions. The limiting factors in these experiments were complications around efficient Monte Carlo sampling, conditional probability estimation, and reliable partial orderings. The first can be entirely resolved with greater computational power. The second requires some care, but may in principle be addressed with flexible methods such as variational autoencoders. The third relies on background knowledge, which may be substantial in some areas, and should in any case grow over time. Improvements in any or all of these areas can help extend the rational Shapley method to larger, more complex cases. 


Another potential challenge to the proposed method is that it is vulnerable to confirmation bias. If users can scan the data in search of attribution vectors that make their preferred outcomes more likely, what’s to stop them from finding such vectors even when do they do not exist? There are three answers to this charge. First, it is entirely possible that the relevant subspace be empty -- that is, no observed or synthetic datapoint meets the threefold criteria specified by $\langle \mathcal{D}_S,  \mathcal{D}_R, \mathcal{Y}_1 \rangle$. 
Such a failure is highly informative, as it demarcates a realm of (im)possibility for the inquiring agent. Thus it is simply false to allege that users will always find what they are looking for. Second, in cases where the relevant subspace is nonempty but sparsely populated, we should expect estimates to be unstable. Any good inference procedure should take such uncertainty into account, making it difficult for an opportunistic user’s desired outcome to pass severe tests on the basis of just a small handful of outlying points. I have not said much about testing here, but it is simple in principle to extend the experiments above with frequentist or Bayesian methods tailored to the particular task at hand \citep{slack_reliable}. Third, there is no guarantee that agents will generally concur on proper values of $\mathcal{D}_S, \mathcal{D}_R$, and $\mathcal{Y}_1$. 
This makes it all the more important to be explicit about which values of each input go into any given explanation. 
Such transparency will make it harder for adversarial agents to game the system and easier for those acting in good faith to come to consensus on particular cases.

\section{Conclusion}\label{sec:conclusion}
The two most popular XAI tools available for local, post-hoc explanations -- feature attributions and counterfactuals -- each have certain advantages. However, they tend to provide different answers in particular cases that can confuse and/or mislead end users. In this paper, I have shown how the two can be synthesized into a single method that preserves the best elements of each. Rational Shapley values preserve and extend the axiomatic guarantees of their classical forebears, giving users the flexibility to compute additive feature attributions in a fast, flexible manner. Marginal, conditional, and interventional value functions were all evaluated, and the applicability of each was compared across different use cases. By formalizing the task in an expected utility framework, I was able to demonstrate how and why individual agents may rationally seek different explanations for the same model predictions. The resulting explanations are concise, intuitive, and thoroughly pragmatic. 


\begin{acks}
This research was supported by ONR grant N62909-19-1-2096. 
\end{acks}

\bibliographystyle{ACM-Reference-Format}
\bibliography{biblio}

\appendix
\section{Appendix}\label{sec:appx}

The following is a proof of Thm.~\ref{thm:rational}, which states that rational Shapley values are the unique additive feature attribution method that satisfies efficiency, linearity, sensitivity, symmetry, and rationality. 

\begin{proof}
    Since uniqueness is already well established for the classical axioms, all that remains is to show that rational Shapley values (a) satisfy the additional rationality axiom and (b) do not violate any of the classical axioms. 
    
    Take (a) first. Assume there exists some $\mathcal{Z}'$ that generates greater expected rewards for an agent than the relevant subspace $\tilde{\mathcal{Z}}$. Then conditioning on the corresponding Shapley values $\bm{\phi}(\bm{z}')$ must lead to an action of greater expected utility than the action recommended by $\bm{\phi}(\tilde{\bm{z}})$. This means that the values for some $\phi_j(\tilde{\bm{z}})$ are either too large or too small. But this can only arise from a misspecification of (i) $\mathcal{Y}_1$, (ii) $\mathcal{D}_S$, or (iii) $\mathcal{D}_R$, which together permit either too much variation in $X_j$ (resulting in inflated values of $|\phi_j(\tilde{\bm{z}})|$), or too little (resulting in deflated values of $|\phi_j(\tilde{\bm{z}})|$). By  Def.~\ref{def:rel_subsp}, any $\tilde{\mathcal{Z}}$ that fails to satisfy one of these criteria is not a relevant subspace. Thus we have a contradiction.
    
    One can verify (b) by confirming that rational Shapley values do not deviate from the classical formulae (see Eqs.~\ref{eq:additive}, \ref{eq:shapley}, \ref{eq:value}). The only difference between the current proposal and more familiar alternatives is their respective methods of specifying baseline expectations $\phi_0$ and reference distributions $\mathcal{D}_R$. Shifting these two parameters changes the interpretation of resulting Shapley vectors, but does not alter their essential character. Indeed, much recent work on Shapley values has been devoted to exploring alternativere reference distributions; for more on the plurality of Shapley values, see \citep{Sundararajan2019, taly2020}. Any procedure that inputs valid values for $\phi_0$ and $\mathcal{D}_R$ and conforms to the characteristic equations enjoys the same properties.
\end{proof}

\end{document}